\documentclass{article}

\PassOptionsToPackage{numbers, sort&compress}{natbib}

\usepackage[final]{neurips_2024_ml4ps}

\usepackage[utf8]{inputenc} 
\usepackage[T1]{fontenc}    
\usepackage{hyperref}       
\usepackage{url}            
\usepackage{booktabs}       
\usepackage{amsfonts}       
\usepackage{nicefrac}       
\usepackage{microtype}      
\usepackage{xcolor}         
\usepackage{amsmath,amssymb,amsfonts}
\usepackage{cleveref}
\usepackage{tikz}
\usepackage{tikz-3dplot}

\usepackage{amsmath, amsfonts, mathtools, amssymb}

\usepackage[color,final]{showkeys}

\newcommand{\cut}[1]{{}}

\makeatletter
\newcommand*{\rom}[1]{\expandafter\@slowromancap\romannumeral #1@}
\makeatother

\newcommand{\cE}{{\mathcal{E}}}

\newcommand{\cN}{{\mathcal{N}}}

\newcommand{\cX}{{\mathcal{X}}}

 \usepackage[bb=boondox]{mathalfa}

\usepackage{xparse}
\DeclareFontFamily{U}{ntxmia}{}
\DeclareFontShape{U}{ntxmia}{m}{it}{<-> ntxmia }{}
\DeclareFontShape{U}{ntxmia}{b}{it}{<-> ntxbmia }{}
\DeclareSymbolFont{lettersA}{U}{ntxmia}{m}{it}
\SetSymbolFont{lettersA}{bold}{U}{ntxmia}{b}{it}

\ExplSyntaxOn
\NewDocumentCommand{\varmathbb}{m}
 {
  \tl_map_inline:nn { #1 }
  {
    \use:c { varbb##1 }
  }
 }
\tl_map_inline:nn { ABCDEFGHIJKLMNOPQRSTUVWXYZ }
 {
  \exp_args:Nc \DeclareMathSymbol{varbb#1}{\mathord}{lettersA}{\int_eval:n { `#1+67 }}
 }
\exp_args:Nc \DeclareMathSymbol{varbbk}{\mathord}{lettersA}{169}
\ExplSyntaxOff

\makeatletter
\DeclareFontFamily{U}{tipa}{}
\DeclareFontShape{U}{tipa}{m}{n}{<->tipa10}{}
\newcommand{\arc@char}{{\usefont{U}{tipa}{m}{n}\symbol{62}}}%

\newcommand{\arc}[1]{\mathpalette\arc@arc{#1}}

\newcommand{\arc@arc}[2]{%
  \sbox0{$\m@th#1#2$}%
  \vbox{
    \hbox{\resizebox{\wd0}{\height}{\arc@char}}
    \nointerlineskip
    \box0
  }%
}
\makeatother

\definecolor{lightgrey}{gray}{0.8}
\definecolor{medgrey}{gray}{0.6}
\definecolor{darkgrey}{gray}{0.4}

\newcommand{\opI}{{\varmathbb{I}}}

\DeclarePairedDelimiter{\norm}{\lVert}{\rVert}

\makeatletter
\newlength{\doublefracgap}
\DeclareRobustCommand{\doublefrac}[2]{%
  \mathinner{\mathpalette\doublefrac@{{#1}{#2}}}%
}
\newcommand{\doublefrac@}[2]{\doublefrac@@#1#2}
\newcommand{\doublefrac@@}[3]{%
  \ooalign{%
    \raisebox{\doublefracgap}{$\m@th#1\frac{#2}{\phantom{#3}}$}\cr
    \raisebox{-\doublefracgap}{$\m@th#1\frac{\phantom{#2}}{#3}$}\cr
  }%
}

\makeatother

\newcommand{\R}{\mathbb{R}}

\newcommand{\diffang}[4]{%
  	\begingroup 
		\def\firstop{#1}%
		\def\secondop{#2}%
		\def\identity{\opI}%
		\ifx\firstop\identity
		\def\firstop{}%
		\fi
		\ifx\secondop\identity
		\def\secondop{}%
		\fi
		\angle (\firstop #3 - \firstop #4, \secondop #3 - \secondop #4)
  	\endgroup
}
\newcommand{\diffnormfrac}[4]{%
  	\begingroup 
		\def\firstop{#1}%
		\def\secondop{#2}%
		\def\identity{\opI}%
		\ifx\firstop\identity
		\def\firstop{}%
		\fi
		\ifx\secondop\identity
		\def\secondop{}%
		\fi
		\frac{\norm{\firstop #3 - \firstop #4}}{\norm{\secondop #3 - \secondop #4}}
  	\endgroup
}

\title{Toward Model-Agnostic Detection of New Physics Using Data-Driven Signal Regions}

\author{%
  Soheun Yi
  \\
  Department of Statistics and Data Science\\
  Carnegie Mellon University\\
  Pittsburgh, PA 15213 \\
  \texttt{soheuny@andrew.cmu.edu} \\
  \And
  John Alison \\
  Department of Physics \\
  Carnegie Mellon University \\
  Pittsburgh, PA 15213 \\
  \texttt{johnalison@cmu.edu} \\
  \AND
  Mikael Kuusela \\
    Department of Statistics and Data Science \\
    Carnegie Mellon University \\
    Pittsburgh, PA 15213 \\
  \texttt{mkuusela@andrew.cmu.edu} \\
}

\begin{document}

\maketitle

\begin{abstract}
In the search for new particles in high-energy physics, it is crucial to select the Signal Region (SR) in such a way that it is enriched with signal events if they are present. While most existing search methods set the region relying on prior domain knowledge, it may be unavailable for a completely novel particle that falls outside the current scope of understanding. We address this issue by proposing a method built upon a model-agnostic but often realistic assumption about the localized topology of the signal events, in which they are concentrated in a certain area of the feature space. Considering the signal component as a localized high-frequency feature, our approach employs the notion of a low-pass filter. We define the SR as an area which is most affected when the observed events are smeared with additive random noise. We overcome challenges in density estimation in the high-dimensional feature space by learning the density ratio of events that potentially include a signal to the complementary observation of events that closely resemble the target events but are free of any signals. By applying our method to simulated $\mathrm{HH} \rightarrow 4b$ events, we demonstrate that the method can efficiently identify a data-driven SR in a high-dimensional feature space in which a high portion of signal events concentrate.
\end{abstract}

\section{Introduction}
\label{s:intro}

High-energy physics is a subfield of physics that studies fundamental particles and their interactions to understand how matter and radiation in the universe are formed. 
At its core, the Standard Model (SM) plays a central role in explaining the fundamental particles and their interactions. 
The remarkable discovery of the Higgs boson by collaborations at the European Organization for Nuclear Research (CERN) \cite{ATLAS2012_observation,CMS2012_observation} and its inclusion in the SM leads to a natural question about the possibility of new particles and interactions beyond the SM.
An important challenge in searching for new physics events is to estimate the background distribution expected from the SM, which is to be compared with the observed data to identify the existence of new signals.
Perhaps the most prominent approaches to background estimation are ABCD and sideband methods, and their variants \cite{ATLAS2012_observation,CMS2012_observation,Bryant2018_search,CDFCollaboration_1991_measurement}. 
At the center of these approaches, one selects the \emph{Signal Region} (SR) where signal events are expected to be concentrated, and the \emph{Control Region} (CR) where signal events are expected to be nearly absent.
Then, the background distribution is estimated from the CR and extrapolated to the SR to compare with the observed data.
Most existing methods invoke knowledge about the expected signals to select the SR and CR.
In detecting the Higgs boson, for example, the SR is configured to include the mass range where the Higgs boson is expected to appear and the CR is configured to exclude the SR \cite{ATLAS2012_observation,CMS2012_observation,Bryant2018_search}.

For a completely new type of particle, however, we are likely to lack specific prior knowledge about the signal events, which has led to extensive recent work on methods for model-agnostic signal searches \cite{LHCO2021_lhc,NachmanShih2020_anomalya,CollinsHoweNachman2018_anomaly,Chakravarti2023,d2019learning,d2021learning, DAgnolo2022}.
In this case, setting up the SR and CR based on existing knowledge about the signal is not feasible.
This motivates us to develop a method to define the SR and CR only with the observed data without relying heavily on prior domain knowledge about the signals.
Rather, we observe that signals tend to be localized in the feature space, 
which is an assumption that has been commonly exploited in certain methods, namely ``bump hunting'' \cite{Choudalakis:2011qn,CollinsHoweNachman2019_extending,CollinsHoweNachman2018_anomaly,BortolatoSmolkovicDillonKamenik2022_bump}.

For instance, the Higgs boson is expected to appear in a particular invariant mass range around $125$ GeV.
Motivated by this observation, we make a minimal assumption that the signal events are localized in the (potentially high-dimensional) feature space and use this to propose a data-driven way of selecting the SR and~CR.


\section{Problem Formulation}

When two protons collide, they form highly unstable particles that decay within an extremely short time frame, making them impossible to detect directly.
Some of these transient particles decay into quarks, which interact with the vacuum to produce collimated sprays of stable particles, referred to as ``jets''. 
Among all possible final states, we focus on events resulting in four jets.

Being able to detect jets efficiently, the Compact Muon Solenoid (CMS) detector at CERN records the momentum and the invariant mass (or energy) of the particles produced in proton-proton collisions. For the momentum of an observed particle, let $(r, \theta, \phi)$ be its representation in spherical coordinates, where $\theta \in [0, \pi)$ is the polar angle and $\phi \in [0, 2\pi)$ is the azimuthal angle.
As visually explained in \Cref{fig:variable-description}, the momentum vector is represented by a three-dimensional vector $(p_T, \eta, \phi)$;
$p_T \coloneqq r \sin \theta$ is the \emph{transverse momentum}, or the magnitude of the momentum projected onto the $xy$-plane, $\eta \coloneqq -\log \tan(\theta / 2)$ is the \emph{pseudorapidity}.
Writing the rest mass of a particle as $m$, the measured features of an observed particle are represented as a four-vector $(p_T, \eta, \phi, m) \in \R^4$.
As we are concerned with collisions resulting in $4$ jets, the corresponding events can be represented with $x = (p_T^i, \eta^i, \phi^i, m^i)_{i=1}^4 \in \R^{16}$, forming a $16$-dimensional feature space $\cX \subset \R^{16}$.

\begin{figure}
	\centering
	\tdplotsetmaincoords{60}{120}
	\begin{tikzpicture}
		[scale=3,
			tdplot_main_coords,
			axis/.style={->,black,thick},
			vector/.style={-stealth,red,very thick},
			vector guide/.style={dashed,red,thick}]
	
		\coordinate (O) at (0,0,0);
		
		\tdplotsetcoord{P}{.8}{55}{60}

		\draw[axis] (0,0,0) -- (1,0,0) node[anchor=north east]{$x$};
		\draw[axis] (0,0,0) -- (0,1,0) node[anchor=north west]{$y$};
		\draw[axis] (0,0,0) -- (0,0,1) node[anchor=south]{$z$};
		
		\draw[vector] (O) -- (P) node[anchor=west]{Momentum $p$};
		
		\draw[vector,black] (O) -- (Pxy) node[anchor=north west]{$p_T$};
		\draw[vector guide,black] (Pxy) -- (P);
		
		\tdplotdrawarc[->]{(O)}{0.2}{0}{60}{anchor=north}{$\phi$};
		\tdplotsetthetaplanecoords{60};
		\tdplotdrawarc[tdplot_rotated_coords,->]{(O)}{0.2}{0}{55}{anchor=south west}{$\theta$};
	\end{tikzpicture}
	\caption{Depiction of the variables used to represent a particle jet. $p_T$ is the transverse momentum, $\phi$ is the azimuthal angle, and $\theta$ is the polar angle. The pseudorapidity is given by $\eta = -\log(\tan(\theta/2))$.}
	\label{fig:variable-description}
\end{figure}

The SM predicts a rare chance of a proton-proton collision producing two Higgs bosons, and in such case, they are most likely to decay into four $b$-quarks which produce specific types of jets called ``$b$-jets''.
This instance of interactions of particles is abbreviated as $\mathrm{HH} \to 4b$ as it incorporates the process of two Higgs bosons decaying into four $b$-jets.
We refer to proton-proton collisions that are tagged to have produced four $b$-jets as $4b$ events, and among them, we are interested in identifying the existence of events corresponding to $\mathrm{HH} \to 4b$.
In the sequel, we refer to these events as the \emph{signal events}, and the rest of the $4b$ events as the \emph{background} $4b$ events.
In addition to the $4b$ events, we have observations of events that are tagged to have produced three $b$-jets; we refer to these events as $3b$ events and make the assumption that the signal we are looking for is not present among the $3b$ events.
We leverage the density ratio of $4b$ events to $3b$ events to select the SR and CR in the $16$-dimensional feature space. We elaborate further on this in \Cref{s:method}.


We have access to $n$ samples of $3b$ and $m$ samples of $4b$ events, denoted by $\{X_{3b}^1, \dots, X_{3b}^n\}$ and $\{X_{4b}^1, \dots, X_{4b}^m\}$, respectively.
Denote these distributions by
\begin{equation}
    X_{3b}^1, \dots, X_{3b}^n \stackrel{\text{iid}}{\sim} P_{3b} 
    , \quad 
    X_{4b}^1, \dots, X_{4b}^m \stackrel{\text{iid}}{\sim} P_{4b} 
    = (1 - \epsilon) B_{4b} + \epsilon S,
\end{equation}
where $B_{4b}$ and $S$ are distributions (or probability measures) of background $4b$ and signal events, respectively.
We call $\epsilon$ the \emph{signal ratio}, which is the fraction of signal events among the observed events.
Since a nonzero $\epsilon$ indicates the existence of a signal in the data, the problem of discovering new physics boils down to testing the following hypotheses:
\begin{equation}\label{eq:sr_ratio_test}
    H_0 \colon \epsilon = 0 \quad \text{vs.} \quad H_1 \colon \epsilon > 0.
\end{equation}

Under the existence of the signal process, our goal is to select the SR, denoted by $\cX_s \subset \cX$, to be rich in signals relative to its size, which can be expressed as amplifying $S(\cX_s)$ compared to $P_{4b}(\cX_s)$.
This is important for improving the power of the test conducted on the signal region since a higher concentration of signal should lead to a greater deviation from the null hypothesis.

\section{Methodology}
\label{s:method}


\textbf{Learning Event Representations.}
An important factor to consider for our collision events is how the four jets are naturally paired into two sets of two jets, according to the intermediate particle from which each jet originates.
In addition, there are symmetries in the feature space (reflection symmetries in $\phi$ and $\eta$, as well as a rotational symmetry in $\phi$) to be considered.
To address the complicated topology in the original feature space posed by these aspects, we exploit the classifier architecture proposed by \citet{Manole2024}, which is designed to handle both the pairing of the jets and the symmetries in the feature space.
Informally speaking, the classifier is trained to separate $3b$ events from $4b$ events and the last layer of the classifier calculates the probability of an event being a $4b$ event as opposed to a $3b$ event from inputs calculated by the previous layers.
We use these inputs to that last layer as the representations of events with the expectation that they exhibit a simpler topology.
We denote the representation of an event $x$ as $\zeta(x)$, or simply $\zeta$ when the context is clear.

\textbf{Finding the Data-Driven Signal Region.}
As mentioned in \Cref{s:intro}, we leverage the assumption that the signal events are localized in the feature space, or the space of representations $\zeta$.
Recall that signal events of our interest are tagged as $4b$ events.
Hence, an area where signals are concentrated would display a higher density of $4b$ events compared to $3b$ events.
In the idealized case of the density ratio of $4b$ events to $3b$ events being uniform across the feature space except at those locations, simply selecting areas near the peak of (or area with the highest value of) the density ratio would be sufficient to identify the SR.

\begin{figure}
    \centering
    \includegraphics[width=0.8\textwidth]{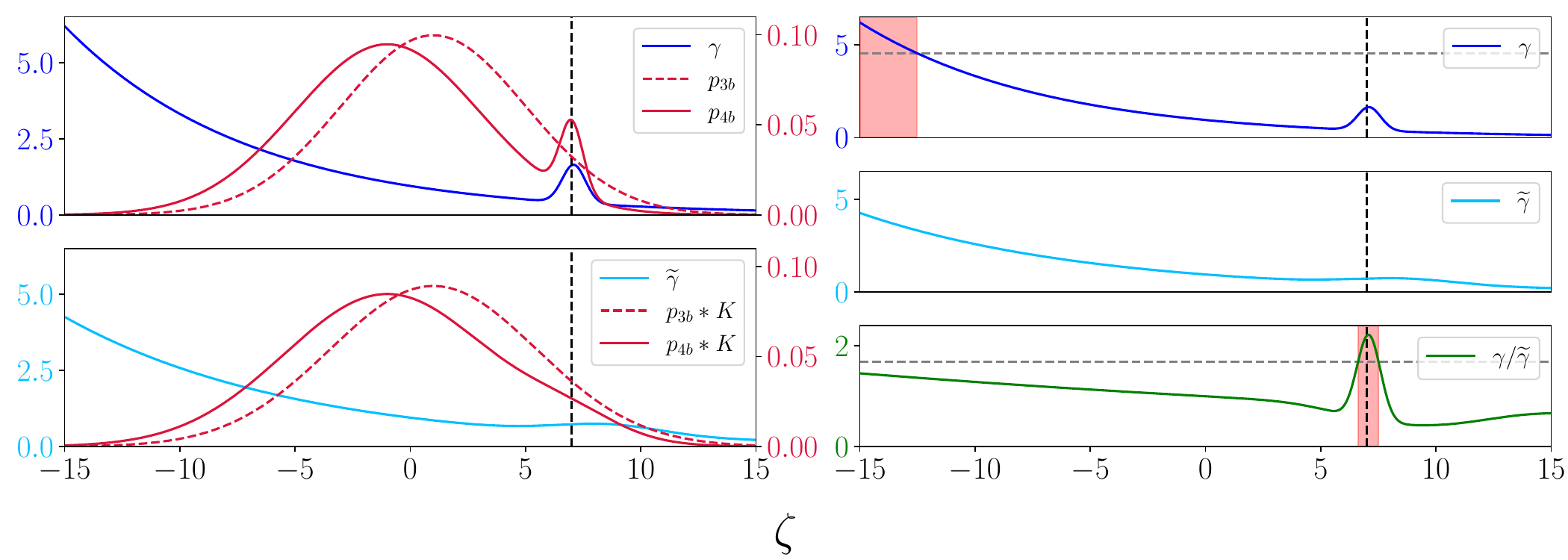}
    \caption{
    $P_{3b} = \cN(1, 4^2)$, $P_{4b} = 0.95 B_{4b} + 0.05 S = 0.95 \cN(-1, 4^2) + 0.05 \cN(7, 0.5^2)$, and $K = \cN(0, 2^2)$. Signal events (with distribution $S$) are concentrated around $\zeta = 7$ (dashed lines).
    Looking at the largest values of $\gamma$ (upper right) does not identify the center of signal events, but the ratio $\gamma / \widetilde{\gamma}$ (lower right) does.}
    \label{fig:smear_to_config_sr}
\end{figure}

Unfortunately, this is not the case in reality. 
It is possible that the density ratio in some areas of the feature space is higher than the ratio in areas where the signal is concentrated, making those regions merely local maxima of the density ratio.
A plot on the upper left of \Cref{fig:smear_to_config_sr} illustrates this issue in a low-dimensional toy example; the center of the signal does not necessarily have the highest value of the density ratio.
To address this issue, we leverage the fact that $3b$ events possess similar (but not identical) kinematics to that of background $4b$ events \citep{CMS2022_search}, which allows us to postulate that the density ratio of background $4b$ events to $3b$ events should only consist of ``low-frequency'' features. 
Therefore, if the density ratio of \emph{all} $4b$ events to $3b$ events possesses a localized ``high-frequency'' feature, it would suggest the presence of a signal in the feature space.
In other words, the density ratio of all $4b$ events to $3b$ events would show a local maximum (or a peak) in the feature space where signal events are concentrated. 

We exploit this topology to find the SR and CR in the high-dimensional feature space.
Motivated by the notion of a low-pass filter, we consider extracting low-frequency components of the $4b$ and $3b$ densities and then comparing the resulting density ratio with the original one to search for high-frequency components.
If we take a convolution kernel (e.g., Gaussian kernel) with adequate bandwidth and convolve each density with it, we can obtain a smoothed version of the density ratio that reflects only low-frequency components in the densities with features corresponding to the signal suppressed.
In mathematical terms, we define the density ratio and the smoothed density ratio as
\begin{align*}
    \gamma(\zeta) \coloneqq \frac{p_{4b}(\zeta)}{p_{3b}(\zeta)}, \quad
    \widetilde{\gamma}(\zeta) \coloneqq \frac{(p_{4b} * K)(\zeta)}{(p_{3b} * K)(\zeta)},
\end{align*}
where $p_{3b}$ (resp. $p_{4b}$) is the density of $3b$ (resp. $4b$) events in the representation space, $*$ denotes the convolution operation, and $K$ is a convolution kernel.
Remarkably, $\widetilde{\gamma}$ can be efficiently estimated by learning a classifier without directly evaluating the convolution operation.
In particular, we can estimate it by training a classifier to distinguish $(Z_{3b} + \cE, 0)$ and $(Z_{4b} + \cE, 1)$, where $Z_{3b}$ and $Z_{4b}$ are the representations of $3b$ and $4b$ events, respectively, and $\cE \sim K$ is random noise.

A toy example shown in \Cref{fig:smear_to_config_sr} illustrates the concept of the smoothed density ratio. One can observe that the center of signal events cannot be identified by selecting a region with the highest value of $\gamma$, but rather by considering the ratio $\gamma / \widetilde{\gamma}$. 
To this end, we define the data-driven SR as
\begin{equation}\label{eq:srcr}
    \cX_s \coloneqq \{ x \mid \gamma(\zeta(x)) / \widetilde{\gamma}(\zeta(x)) \geq \tau_s \}.
\end{equation} 
The threshold $\tau_s$ is specifically designed to control the size of the SR as measured by the proportion of $4b$ events within those regions.

\section{Results}

In training classifiers to estimate density ratios, we used $3b$ event samples of size $n \in \{10^5, 10^6\}$ and the same size of $4b$ events. 
We used $75\%$ and $6.25\%$ of all the samples to estimate $\gamma$ and $\widetilde{\gamma}$, respectively \footnote[1]{The rest of the samples were left for the hypothesis testing.}.
To generate the training dataset for the smoothed density ratio $\widetilde{\gamma}$, Gaussian noise was added to the representations of each event, with noise scale in each dimension set to $\eta \in \{0.01, 0.1, 1\}$ times the length of the range of the corresponding representation.

\begin{figure}
    \centering
    \includegraphics[width=0.9\textwidth]{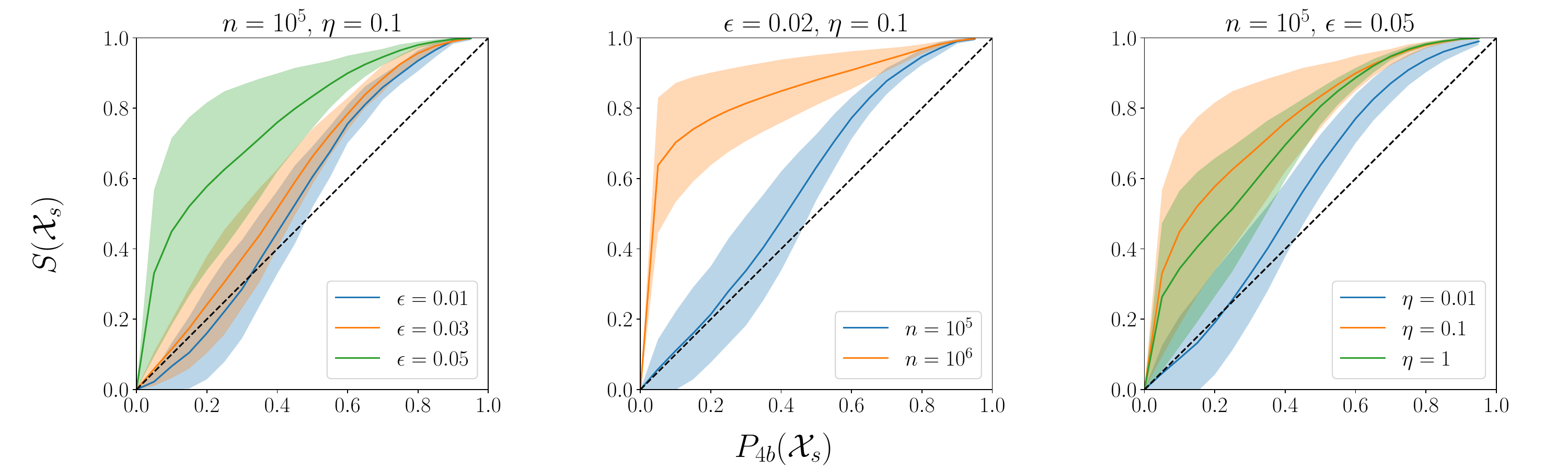}
    \caption{Concentration of signal events in the SR for the $\mathrm{HH} \to 4b$ data. $x$ axis represents the proportion of $4b$ events in the SR (measured by $P_{4b}(\cX_s)$) and $y$ axis represents the proportion of signal events in the SR (measured by $S(\cX_s)$). 
    As defined previously, $n$, $\epsilon$, and $\eta$ are the number of $3b$ events, the signal ratio, and the scale of the convolution kernel, respectively.
    Bold lines and shaded areas represent (mean) $\pm$ (standard deviation) measured through $10$ repeated experiments.
    }
    \label{fig:SR_efficiency}
\end{figure}

Recall that our goal is to enrich the SR (i.e., $\cX_s$) with the signal, that is, to have a high portion of signal events in the SR (i.e., $S(\cX_s)$) compared to the size of the SR measured by the portion of $4b$ events in the SR (i.e., $P_{4b}(\cX_s)$).
In an ideal case, the curve should rapidly increase to $1$ as the size of SR increases.
As \Cref{fig:SR_efficiency} suggests, our method can find an SR that shows such behavior, especially when the signal ratio is high (left) or the sample size is large (center). 
Furthermore, the plot on the right shows that setting $\eta = 0.1$ gives the best performance, demonstrating the importance of choosing a convolution kernel whose size is aligned with the localized signal. 

\begin{figure}
    \centering
    \includegraphics[width=0.42\textwidth]{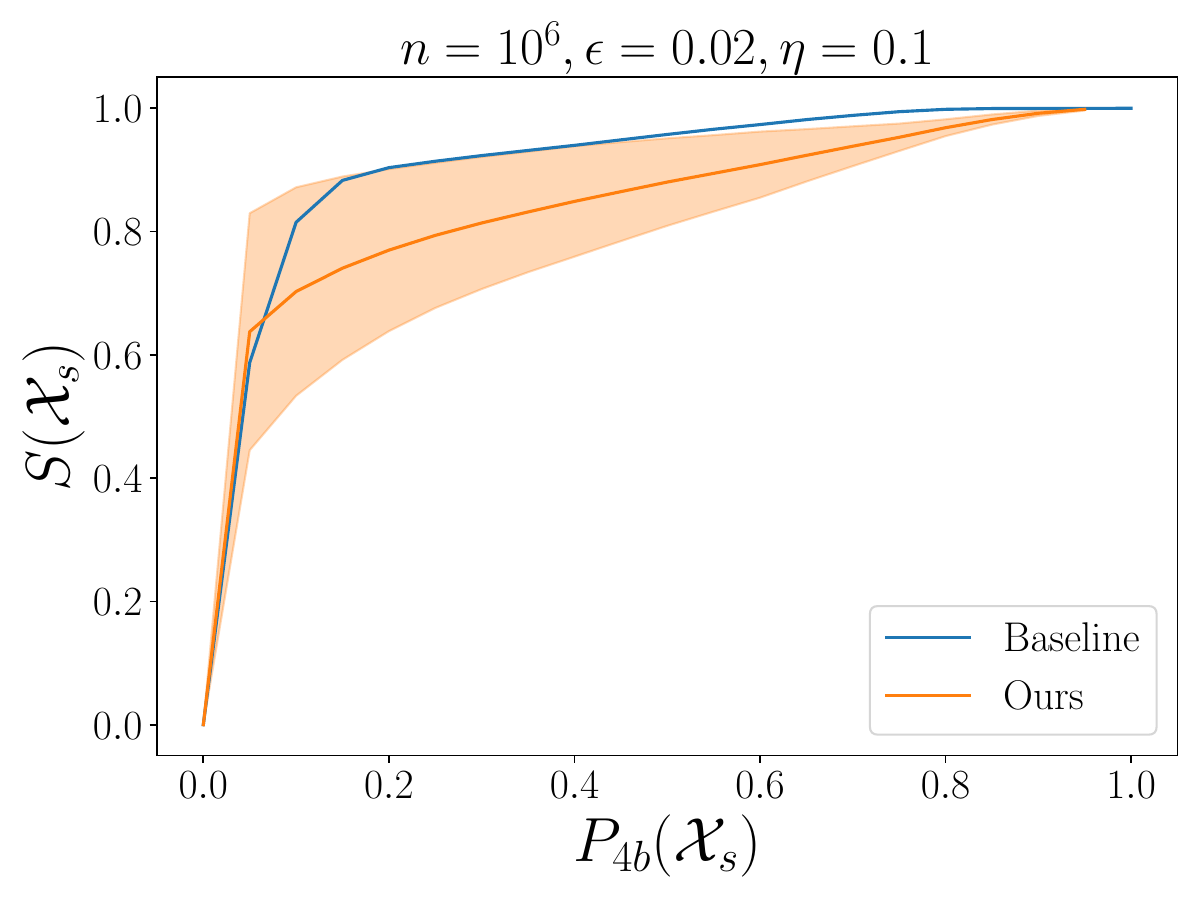}
    \caption{
    Comparison with the baseline SR provided by \citet{Bryant2018_search}. 
    Bold lines represent the average measured through $10$ repeated experiments, and the shaded region represents the standard deviation of our method (the baseline has an ignorable variance, which is hence omitted).
    While the baseline, which has access to a priori knowledge of the location of the signal, shows better performance when $P_{4b}(\cX_s)$ is large, the methods have comparable performance when $P_{4b}(\cX_s)$ is small.}
    \label{fig:SR_comparison}
\end{figure}

\Cref{fig:SR_comparison} compares our method to a baseline SR proposed by \citet{Bryant2018_search}. 
The baseline SR is designed to capture events with invariant masses of both intermediate particles close to that of the Higgs boson.
We emphasize that the baseline is aware of particular domain knowledge about the signal, which we do not require.
While \Cref{fig:SR_comparison} shows a performance gap between the two methods, our model-agnostic SR remains competitive, or even slightly better when the size of SR ($P_{4b}(\cX_s)$) is small.
In hypothesis testing, it is crucial to limit the size of the SR to maintain a high signal-to-background ratio in the region. 
From this perspective, our method achieves signal concentration that is comparable with the baseline while requiring weaker domain-specific knowledge about the signal.

\section{Discussion}

The plot on the center of \Cref{fig:SR_efficiency} shows that an increase in the sample size leads to a better concentration of signals in the SR, even when the signal ratio is fixed to a low value. 
This suggests that our method has sensitivity for small signals as long as the sample size is large enough.
We also remark on the importance of choosing a convolution kernel introducing an optimal amount of noise to the representations of events in learning the smoothed density ratio $\widetilde{\gamma}$.
If we choose the noise with too large a scale compared to the size of a signal peak, the density ratio would be overly smoothed, affecting the low-frequency features that we intend to preserve in $\widetilde{\gamma}$. 
Conversely, using too small noise would render $\widetilde{\gamma}$ indistinguishable from the original density ratio $\gamma$. 
In both cases, we fail to capture the peak we are searching for. 

The current scope of our work does not include the estimation of the background $4b$ distribution and the hypothesis testing (\Cref{eq:sr_ratio_test}) to determine the presence of the signals in the SR.
We plan to extend our work to include these steps in future work.

\begin{ack}
This work was supported in part by NSF grant DMS-2053804. We are grateful to the members of the STAMPS@CMU Research Center for helpful discussions and feedback on this work.
\end{ack}

\bibliographystyle{plainnat}
\bibliography{references}

\end{document}